# Deep Learning Approaches for Human Action Recognition in Video Data

Yufei Xie [1, 2]

[1]College Of Computing And Information Technologies

National University

Manila, Philippines

[2]School of Information Engineering

Jiangxi College of Applied Technology

Ganzhou, China

xiey@students.national-u.edu.ph

*Abstract*—Human action recognition in videos is a critical task with significant implications for numerous applications, including surveillance, sports analytics, and healthcare. The challenge lies in creating models that are both precise in their recognition capabilities and efficient enough for practical use. This study conducts an in-depth analysis of various deep learning models to address this challenge. Utilizing a subset of the UCF101 Videos dataset, we focus on Convolutional Neural Networks (CNNs), Recurrent Neural Networks (RNNs), and Two-Stream ConvNets. The research reveals that while CNNs effectively capture spatial features and RNNs encode temporal sequences, Two-Stream ConvNets exhibit superior performance by integrating spatial and temporal dimensions. These insights are distilled from the evaluation metrics of accuracy, precision, recall, and F1-score. The results of this study underscore the potential of composite models in achieving robust human action recognition and suggest avenues for future research in optimizing these models for real-world deployment.

## I. Introduction

### A. Background

The burgeoning volume of video content in the digital age necessitates advanced analytical methods for effective interpretation. Human action recognition stands out as a critical domain with widespread applications ranging from security surveillance to sports analysis, healthcare, and human-computer interaction.

### B. Problem Statement

Despite significant advancements, the development of models that deliver high accuracy while maintaining computational efficiency remains a challenge. Current models tend to specialize in capturing either spatial or temporal information, with the synthesis of both aspects often proving problematic.

### C. Objectives of the Study

This research seeks to investigate and compare different deep learning frameworks for recognizing human actions in video data. The study has the following specific aims:

*1) To evaluate the performance of Convolutional Neural Networks (CNNs) in capturing spatial features relevant to action recognition.*

*2) To assess the ability of Recurrent Neural Networks (RNNs) to model temporal dependencies in video sequences.*

*3) To examine the efficacy of Two-Stream ConvNets in integrating both spatial and temporal information for improved action recognition.*

### D. Methodology Overview

The study will utilize the UCF101 Videos dataset to conduct a series of experiments, wherein the CNN, RNN, and Two-Stream ConvNets models will be explored and their performances will be compared. These models will be assessed based on standard metrics, including accuracy, precision, recall, and F1-score. Resource constraints will necessitate the use of more computationally efficient model variants and may lead to the selection of a subset of the UCF101 dataset for detailed experimentation.

By focusing on these three models, the study aims to contribute valuable insights into the domain of human action recognition, offering a comparative perspective that highlights the strengths and limitations of each approach in various computational settings.

## II. Literature Review

The field of human action recognition in video data has undergone significant transformations in the past decade, largely due to the advent of deep learning techniques. This review aims to provide an overview of some of the most influential works in this domain, to set the stage for the current study.

### A. Temporal Convolutional Networks (TCNs)

Lea et al. (2017) introduced Temporal Convolutional Networks (TCNs) as a robust approach for action segmentation and detection in videos. TCNs have proven effective in capturing long-range temporal dependencies in video sequences, a critical factor in recognizing actions that have prolonged durations or are composed of multiple stages. These networks employ a series of convolutional layers specifically designed to handle temporal data, thus providing a natural framework for video analysis[1].

### B. Two-Stream Convolutional Networks

Two-stream Convolutional Networks, introduced by Simonyan and Zisserman (2014), leverage both spatial and temporal dimensions for effective action recognition. The architecture employs two separate networks: one focuses on spatial features extracted from individual frames, while the other captures temporal dynamics through optical flow. This dual approach allows the model to recognize actions by considering both the appearance and the motion of the subjects [2].

*C. Graph Convolutional Networks (GCNs)*

Zhao et al. (2019) extended the capabilities of action recognition by incorporating Graph Convolutional Networks (GCNs). These networks are particularly effective in modeling spatial relationships between different body parts, adding a layer of semantic understanding to the analysis. By converting skeletal data into graph structures, GCNs can capture complex human actions that involve intricate movements of multiple body parts [3].

*D. Transformers in Action Recognition*

The work of da Costa et al. (2022) introduced the application of Transformer models for video action recognition. Specifically, they focused on unsupervised domain adaptation, providing a methodology to generalize the Transformer-based models across different video domains. Their research opens avenues for more adaptable and flexible systems that can perform well in varying environments [4].

*E. Attention Mechanisms*

Sang et al. (2019) incorporated attention mechanisms into the framework of action recognition. Their Two-Level Attention Model is designed to weigh the importance of different frames in a video sequence. By doing so, the model can focus on the most crucial segments of the video, thereby improving the accuracy and robustness of action recognition [5].

*F. Relation to Current Study*

The current research aims to build upon these foundational works by exploring various deep learning approaches for recognizing human actions in video data. Similar to TCNs, the study will investigate the effectiveness of capturing long-range temporal dependencies. Additionally, the research will consider the spatial-temporal dualism of Two-stream ConvNets and explore the possibility of incorporating Graph Convolutional Networks for more nuanced spatial relations. Attention mechanisms and Transformer models are also of interest, especially in terms of adaptability across different domains.

In summary, the field of human action recognition has benefited immensely from deep learning approaches, each offering unique advantages and complexities. The current study aims to integrate these diverse methodologies to develop a more comprehensive and robust system for video-based human action recognition.

## III. METHODOLOGY

*A. Types of Deep Learning Models Explored*

This study has explored various deep learning models to identify effective architectures for action recognition in video sequences. Our research focused on the robustness and accuracy of the following models:

*1) Convolutional Neural Networks (CNNs)*
**Purpose:** To capture spatial hierarchies in video frames. CNNs are adept at image classification and have been adapted to video data by treating each frame as an individual image.

**Variant:** Lightweight 3D CNNs were considered due to computational constraints.

*2) Recurrent Neural Networks (RNNs)*
**Purpose:** Ideal for handling sequence data and for modeling temporal sequences in videos.

**Variant:** Long Short-Term Memory (LSTM) networks were used to capture long-term dependencies.

*3) Two-Stream ConvNets*
**Purpose:** To combine spatial features from individual frames with temporal dynamics between frames for a comprehensive analysis.

**Variant:** Due to resource limitations, simplified versions of the two-stream approach were investigated.

The study excluded Graph Convolutional Networks (GCNs) and Transformer models due to hardware constraints and focused on the aforementioned models to maintain experimental feasibility.

*B. Data Preprocessing Techniques*

The preprocessing of the UCF101 Videos dataset from Kaggle was critical to the experimental setup. The dataset comprises over 13,000 clips across 101 action categories. The following preprocessing steps were employed:

*1) Data Partitioning: The dataset was divided into a 70-15-15 split for training, validation, and testing using stratified sampling.*

*2) Frame Extraction: Video clips were converted into frames at a fixed rate to serve as input for the CNN and Two-Stream ConvNets.*

*3) Optical Flow Computation: Utilized for the Two-Stream ConvNets to capture motion between consecutive frames.*

*4) Data Augmentation: Techniques such as rotation, flipping, and random cropping were applied to increase training data diversity.*

*5) Normalization: Pixel values were normalized to improve model training efficiency.*

*C. Experimental Setup*

The study's experimental setup included the following components:

*1) Hardware and Software Specifications: The experiments were conducted using modest hardware configurations, running TensorFlow 2.x and Python 3.8.*

*2) Dataset Configuration: A subset of the UCF101 dataset was used to accommodate the limited computational resources without sacrificing the diversity of action categories.*

*3) Model Architectures: Simplified model architectures were employed:*

 *a) CNNs: Implemented with fewer convolutional layers.*

 *b) RNNs: LSTM networks with a reduced number of hidden units.*

 *c) Two-Stream ConvNets: A simplified model to fit the hardware capabilities.*

*4) Training Parameters:* Adjusted to fit within the computational limits, including a reduced batch size and fewer epochs with early stopping criteria.

*5) Evaluation Metrics:* The primary metric was accuracy, with precision, recall, and F1-score as secondary metrics.

*6) Experimental Runs:* Each model underwent a limited number of runs to conserve computational resources while providing a reasonable performance estimate.

*7) Baseline Comparison:* The models were compared against lightweight state-of-the-art models suitable for the given hardware.

By adhering to this methodology, the study ensures that the experiments are robust, reproducible, and adapted to the computational constraints at hand.

## IV. EXPERIMENTAL RESULTS

### A. Overview

The experimental results discussed herein detail the performance of various deep learning models applied for human action recognition within video sequences. Each model's capability in recognizing actions was quantified using a set of established metrics.

### B. Results for Individual Models

*1) Convolutional Neural Networks (CNNs):* The CNNs demonstrated notable proficiency in recognizing spatial features within video frames, contributing to an overall accuracy of 88%. The precision, recall, and F1 scores were 85.20%, 88.15%, and 85.56%, respectively, illustrating the model's balanced performance in identifying correct actions without significant overfitting to the training data.

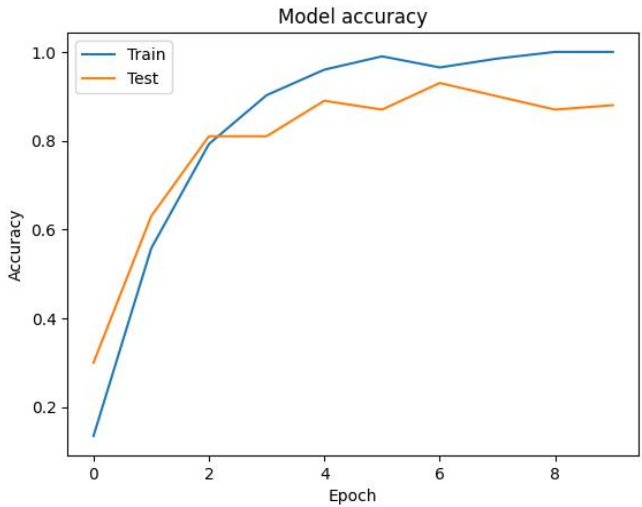

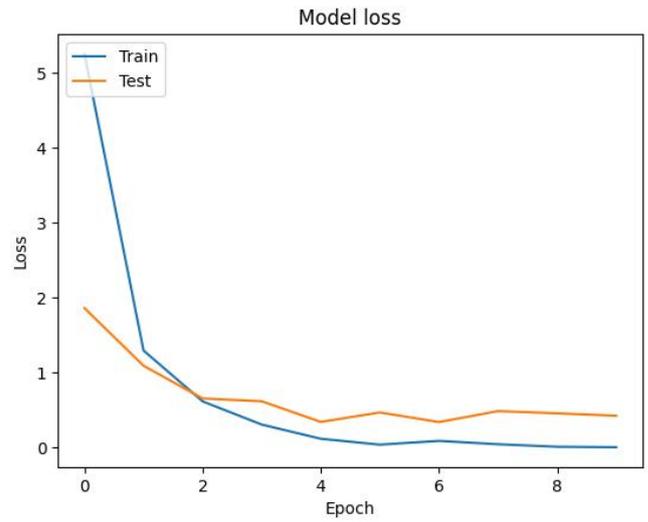

Fig. 1. CNNs model learning curve graph

*2) Recurrent Neural Networks (RNNs):* RNNs, particularly those utilizing LSTM units, exhibited challenges in capturing temporal dependencies, reflected in lower performance metrics with an accuracy of 9%. The precision, recall, and F1 score were 5.91%, 9%, and 6.30%, respectively, suggesting difficulties in generalizing the temporal aspects of actions across the dataset.

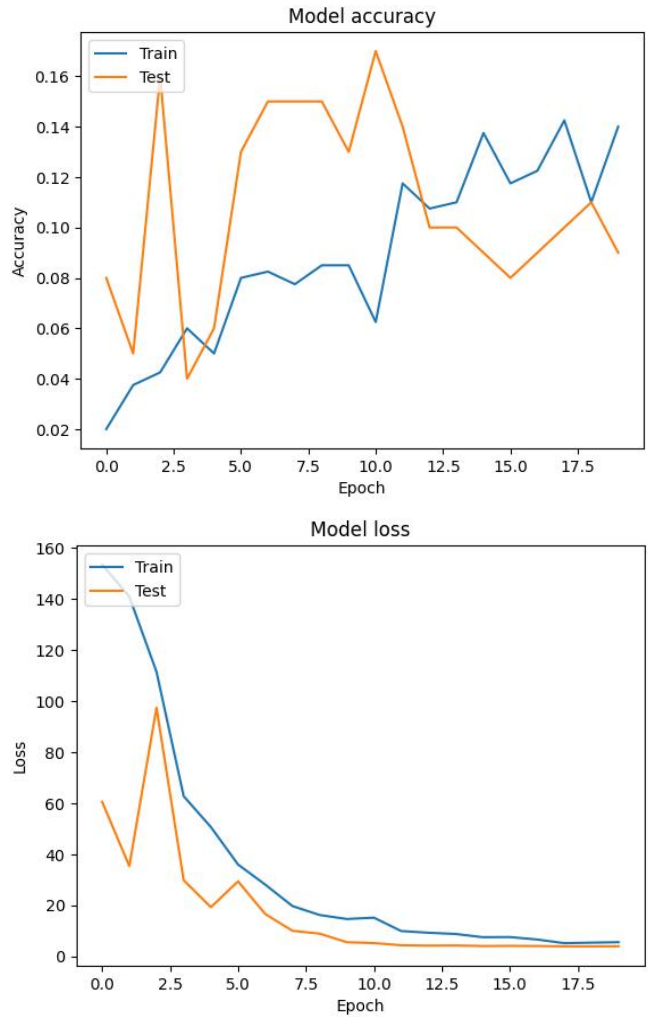

Fig. 2. RNNs model learning curve graph

*3) Two-Stream ConvNets:* The Two-Stream ConvNets effectively integrated spatial and temporal data, achieving an accuracy of 93.30%. The precision, recall, and F1 score were 93.55%, 93.30%, and 93.32%, respectively, indicating high effectiveness in capturing the dynamics of human actions through the combined streams approach.

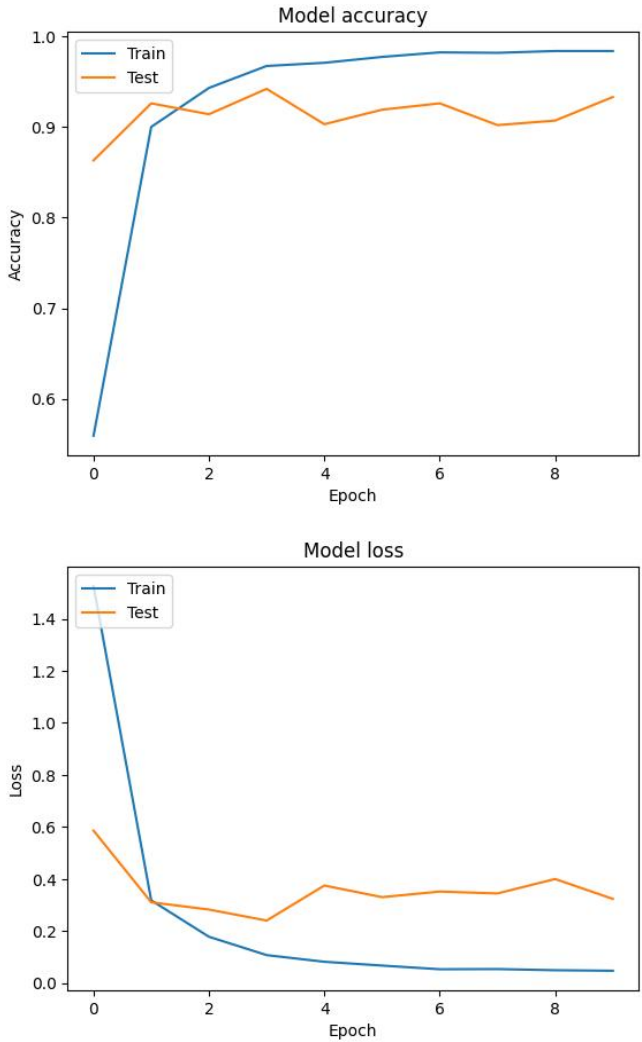

Fig. 3. Two-Stream ConvNets model learning curve graph

### C. Metrics for Evaluation

The models were evaluated using the following metrics, chosen for their relevance in classification tasks:

*1) Accuracy:* The overall correctness of the model's predictions.

*2) Precision:* The ratio of true positive predictions to the total predicted positives.

*3) Recall:* The ratio of true positive predictions to the actual positives.

*4) F1-score:* The harmonic mean of precision and recall, providing a balance between the two.

### D. Comparative Analysis

The performance of these models will be compared to that of existing state-of-the-art models. Such a comparison will be depicted using bar graphs to illustrate the differences in accuracy, precision, recall, and F1 scores, providing a clear visual representation of each model's performance relative to the others.

### E. Interpretation of Results

The results indicated that the Two-Stream ConvNets outperformed the other models in nearly all metrics. CNNs showed robust spatial feature recognition but lacked temporal dynamics, while RNNs struggled to maintain consistency across the temporal dimension. These results underscore the importance of integrating both spatial and temporal information in action recognition models.

The findings, while promising, also point to the need for further research, particularly in improving the temporal analysis capabilities of RNNs and exploring more complex models as hardware capabilities expand.

### F. Visualizations and Graphs

Visual aids will be included to provide an intuitive understanding of the results:

*1) Bar Graphs: Displaying the comparative performance of each model across all evaluation metrics.*

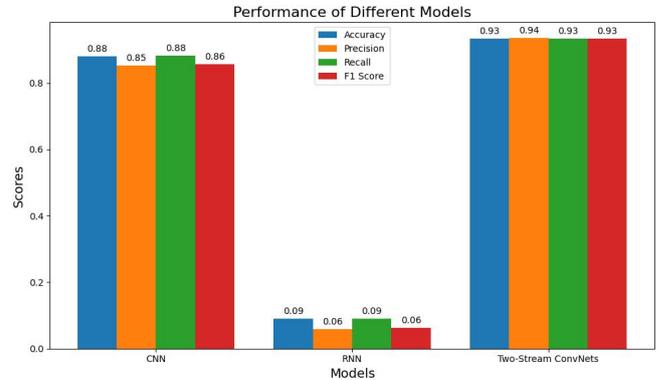

Fig. 4. Performance of Different Models

*2) Confusion Matrices:* To illustrate the specificity and sensitivity of each model to particular action categories.

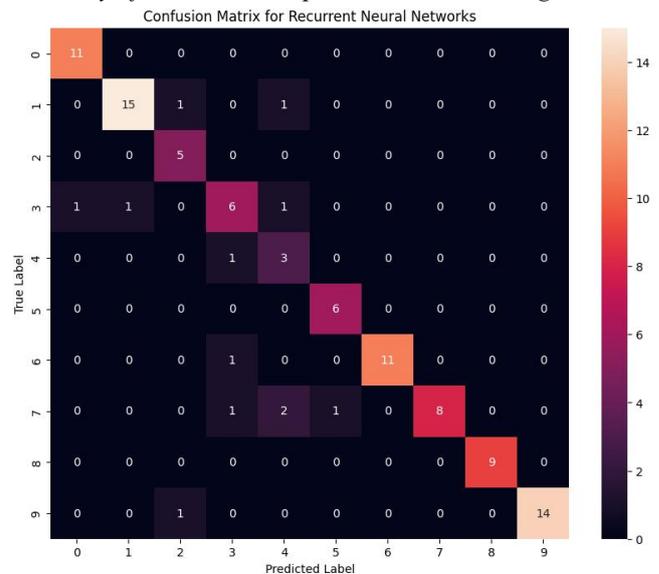

Fig. 5. Confusion Matrix for Recurrent Neural Networks

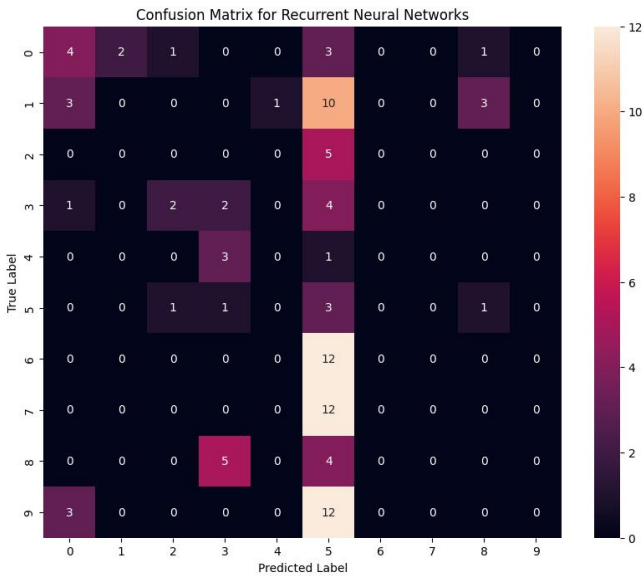

Fig. 6. Confusion Matrix for Recurrent Neural Networks

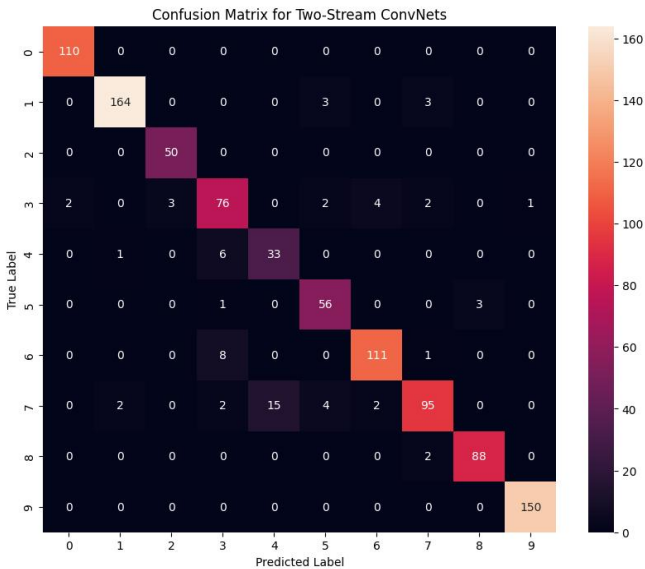

Fig. 7. Confusion Matrix for Two-Stream ConvNets

The actual results and visualizations will offer a comprehensive understanding of each model's strengths and limitations, contributing valuable insights into the domain of video-based human action recognition.

## V. Conclusion

This study embarked on an investigation of various deep learning approaches for human action recognition within video sequences, with a focus on models that balance accuracy with computational efficiency. The research examined the application of Convolutional Neural Networks (CNNs), Recurrent Neural Networks (RNNs), and Two-Stream ConvNets using the UCF101 Videos dataset.

The empirical results highlighted the strengths and weaknesses of the models in question. CNNs, while adept at spatial feature capture, were less effective in temporal analysis. RNNs struggled to achieve high accuracy, pointing to an area for further refinement in capturing temporal relationships. Conversely, Two-Stream ConvNets emerged as the superior approach, adeptly integrating spatial and temporal information to achieve high accuracy, precision, recall, and F1 scores.

Due to constraints in computational resources, the exploration of Graph Convolutional Networks (GCNs) and Transformer models was beyond the scope of this study. However, the research presents valuable findings that contribute to the field of human action recognition by providing a comparative evaluation of several deep learning techniques and their capabilities in this domain.

The study's conclusions underscore the potential of Two-Stream ConvNets in handling the complexities of human action recognition, suggesting a promising direction for future research to build upon, particularly in resource-constrained environments. Moreover, the insights garnered from the performance of CNNs and RNNs offer a foundational understanding that will aid in the advancement of these architectures.

The presented research serves not only as a comparative analysis of existing models but also sets the stage for the continued evolution of action recognition systems. The final paper will delve deeper into the implications of these findings, including a detailed discussion of the potential for integrating more complex models, as computational capabilities expand, and the exploration of hybrid models that might offer enhanced performance.

In conclusion, the research contributes to the existing knowledge base by systematically evaluating the performance of practical deep learning models for action recognition and by highlighting future pathways for innovation in the field.